%% file: main.tex
\documentclass[runningheads]{llncs}
\usepackage[T1]{fontenc}
\usepackage{multirow}
\usepackage[utf8x]{inputenc}
\usepackage[vietnamese, english]{babel}
\usepackage{mathptmx}
\usepackage{array}
\usepackage{booktabs}
\usepackage{graphicx}
\usepackage{caption}
\usepackage{booktabs}
\usepackage{microtype}
\usepackage{longtable}
\usepackage{lmodern}

\begin{document}
\title{SPBERTQA: A Two-Stage Question Answering System Based on Sentence Transformers for Medical Texts}
%
%

\author{Nhung Thi-Hong Nguyen\inst{1,2} \and
Phuong Phan-Dieu Ha\inst{1,2} \and
Luan Thanh Nguyen\inst{1,2}\thanks{Corresponding Author} \and Kiet Van Nguyen\inst{1,2} \and  Ngan Luu-Thuy Nguyen\inst{1,2}}
\authorrunning{Nguyen et al.}
\titlerunning{A Two-Stage Question Answering System Based on Sentence Transformers}
%
\institute{University of Information Technology, Ho Chi Minh City, Vietnam \and
Vietnam National University Ho Chi Minh City, Vietnam\\ 
\email{\{18521218,18521268\}@gm.uit.edu.vn} 
\\
\email{\{luannt,kietnv,ngannlt\}@uit.edu.vn}}
\maketitle              
\begin{abstract}
Question answering (QA) systems have gained explosive attention in recent years. However, QA tasks in Vietnamese do not have many datasets. Significantly, there is mostly no dataset in the medical domain. Therefore, we built a \textbf{Vi}etnamese \textbf{Health}care \textbf{Q}uestion \textbf{A}nswering dataset (ViHealthQA), including 10,015 question-answer passage pairs for this task, in which questions from health-interested users were asked on prestigious health websites and answers from highly qualified experts. This paper proposes a two-stage QA system based on Sentence-BERT (SBERT) using multiple negatives ranking (MNR) loss combined with BM25. Then, we conduct diverse experiments with many bag-of-words models to assess our system's performance. With the obtained results, this system achieves better performance than traditional methods.

\end{abstract}

\input{sections/1_introduction}
\input{sections/2_relatedwork}
\input{sections/3_taskdescription}
\input{sections/4_dataset}

\input{sections/5_methodologies}

\input{sections/6_experiments}
\input{sections/7_resultsanddiscussion}

\input{sections/8_conclusion}
\setlength{\tabcolsep}{4pt}

\section*{Acknowledgement}
Luan Thanh Nguyen was funded by Vingroup JSC and supported by the Master Scholarship Programme of Vingroup Innovation Foundation (VINIF), Vingroup Big Data Institute (VinBigdata), VINIF.2021.ThS.41.

\bibliographystyle{plain}
\bibliography{ref} 
\end{document}

%% file: sections/1_introduction.tex
\section{Introduction}
Today, many websites have QA forums, where users can post their questions and answer other users’ questions. However, they usually take time to wait for responses. Moreover, data for question answering has become enormous, which means new questions inevitably have duplicate meanings from the questions in the database. In order to reduce latency and effort, QA systems based on information retrieval (IR) retrieving a good answer from the answer collection is essential. QA relies on open domain datasets such as texts on the web or closed domain datasets such as collections of medical papers like PubMed \cite{PubMedQA} to find relevant passages. Moreover, in the COVID-19 pandemic, people care more about their health, and the number of questions posted on health forums has increased rapidly. Therefore, QA in the medical domain plays an important role. 
Lexical gaps between queries and relevant documents that occur when both use different words to describe similar contents have been a significant issue. Table \ref{example} shows a typical example of this issue in our dataset. Previous studies applied word embeddings to estimate semantic similarity between texts to solve \cite{Ye}. Various research studies approached deep neural networks and BERT to extract semantically meaningful texts \cite{laskar2020contextualized}. Primarily, SBERT has recently achieved state-of-the-art performance on several tasks, including retrieval tasks \cite{Henderson}. This paper focuses on exploring fine-tuned SBERT models with MNR.

We contribute: (1) Introduce a ViHealthQA dataset containing 10,015 pairs in the medical domain. (2) Propose two-stage QA system based on SBERT with MNR loss. (3) Perform multiple experiments, including traditional models such as BM25, TF-IDF cosine similarity, and Language Model to compare our system.
\vspace*{-20pt}
\begin{table}
\centering
\caption{A typical example of Lexical gaps in ViHealthQA dataset.}
\label{example}
\selectlanguage{vietnamese}
\begin{tabular}{cp{0.8\textwidth}}
  \toprule

  \textbf{ID} & 392 \\
  \midrule
  \textbf{Question}& Tôi bị dị ứng thuốc kháng sinh và dị ứng khi ăn thịt cua đồng. Trường hợp của tôi có được tiêm vaccine phòng Covid-19 không?
  
(\emph{I am allergic to antibiotics and eating crab meat. Can my case be vaccinated against Covid-19?})
 \\
  \midrule
  \textbf{Answer passage}& Trường hợp của anh theo hướng dẫn của Bộ Y tế là thuộc đối tượng cần cẩn trọng khi tiêm vaccine Covid-19 và tiêm tại bệnh viện hoặc cơ sở y tế có đầy đủ năng lực cấp cứu ban đầu.
  
(\emph{According to the guidance of the Ministry of Health, your case is one of the subjects that need to be careful when injecting the Covid-19 vaccine and injecting it at a hospital or medical facility with total initial first aid capacity.})
\\
  \bottomrule
\end{tabular}

\end{table}
\vspace*{-20pt}

%% file: sections/2_relatedwork.tex
\section{Related work}
In early-stage works of QA retrieval, several studies \cite{Salton} presented sparse vector models. Using unigram word counts, these models map queries and documents to vectors having many 0 values and rank the similarity values to extract potential documents. In 2008, Manning et al. \cite{Manning} did many experiments to gain a deeper understanding of the role of vectors, including how to compare queries with documents. Moreover, many researchers \cite{Gery,Robertson} pay attention to BM25 methods in IR tasks.

IR methods with sparse vectors have a significant drawback: lexical gap challenges. The solution to this problem is using dense embedding to represent queries and documents. This idea was proposed early with the LSI approach \cite{Deerwester}. However, the most well-known model is BERT. BERT applied encoders to compute embeddings for the queries and the documents. Liu et al. \cite{liu} installed the final mean pooling layer and then calculated similarity values between outputs. Instead, Karpukhin et al. \cite{karpukhin} used the initial CLS token. Many studies \cite{laskar,Lee} applied BERT and reached significant results. Significantly, SBERT \cite{Reimers} uses Siamese and triplet network structures to represent semantically meaningful sentence embeddings. Multiple research approaches have approached SBERT for Semantic Textual Similarity (STS) and Natural Language Inference (NLI) benchmarks. In 2021, Ha et al. \cite{Ha} utilized SBERT to find similar questions in community question answering. They did several experiments on SBERT with multiple losses, including MNR loss.

Because of our task in the medical domain, we reviewed some related corpus. For example, CliCR \cite{CliCR} comprises around 100,000 gap-filling queries based on clinical case reports, and MedQA \cite{Zhang} includes answers for real-world multiple-choice questions. In Vietnam, Nguyen et al., 2021 \cite{van2020new} published ViNewsQA, including 22,057 human-generated question-answer pairs. This dataset supports machine reading comprehension tasks. 

%% file: sections/3_taskdescription.tex
\section{Task description}
There are $n$ question-answer passage pairs in the database. We have a collection of questions ${Q = \{q_1, q_2, ...,q_n \}}$ and a collection of answer passages ${A = \{a_1,a_2,...,a_n\}}$. Our task is creating models with question $i$ ${ (q_i)}$ belongs to collection $Q$ ${ (q_i \in Q)}$ can retrieve precise answer passage  ${a_i}$ ${ (a_i \in A)}$.

%% file: sections/4_dataset.tex
\section{Dataset}
\subsection{Dataset characteristics}
 We release ViHealthQA, a novel Vietnamese dataset for question answering and information retrieval, including 10,015 question-answer passage pairs. We collect data from Vinmec\footnote{https://www.vinmec.com/} and VnExpress\footnote{https://vnexpress.net/} websites by using the BeautifulSoup\footnote{https://pypi.org/project/beautifulsoup4/} library. These ones are forums where users ask health-related questions answered by qualified doctors. The dataset consists of 4 features: index, question, answer passage, and link.   

\subsection{Overall statistics} 
After the collecting data phase, we divide our dataset into train, dev, and test sets. In particular, there are 7,009 pairs in Train, 993 pairs in Dev, and 2,013 pairs in Test (Table \ref{tab:ques_len}).

According to Table \ref{tab:len}, most of the answer passages are in the range of 101 – 300 words (34.1\%), the second ratio is the number of answer passages with 301 – 500 words (31.13\%), followed by 501 - 700 words (15.88\%), and 701 - 1000 words (9.98\%). Longer answer passages (over 1000 words) comprise a small proportion (above 7.58\%).

\begin{table}
\begin{minipage}{.45\linewidth}
\caption{Statistics of ViHealthQA dataset.}
\label{tab:ques_len}
\begin{tabular}{lc}
\hline
  \textbf{ViHealthQA}     &    \textbf{Value}           \\ \hline

Train &  7,009\\
Dev &  993\\
Test & 2,013\\
Average length answer & 495.33\\
Average length question & 103.87\\
Vocabulary (word) & 18,271\\
Average number of sentences
 & 3.95\\
\hline
\end{tabular}

\end{minipage}
\hfill
\begin{minipage}{.45\linewidth}
\vspace*{-10pt}

\caption{Distribution of the answer passage length (\%).}
\label{tab:len}

\begin{tabular}{lllll}

\hline
\multicolumn{1}{c}{\multirow{2}{*}{\textbf{Length}}} & \multicolumn{4}{c}{\textbf{Answer passage}} \\ \cline{2-5} 
\multicolumn{1}{c}{} & \multicolumn{1}{c}{\textbf{Train}} & \multicolumn{1}{c}{\textbf{Val}} & \multicolumn{1}{c}{\textbf{Test}} & \multicolumn{1}{c}{\textbf{All}} \\ \hline
\textless{} 100 & 1.24 & 1.31 & 1.64 & 1.33 \\
101 – 300 & 34.46 & 34.34 & 32.89 & 34.1 \\
301 – 500 & 31.13 & 30.72 & 31.25 & 31.13 \\
501 – 700 & 15.99 & 16.31 & 15.2 & 15.88\\
701 – 1000 & 9.8 & 8.66 &11.23 & 9.98\\
\textgreater{} 1000 &7.38 & 8.66 & 7.8 & 7.58 \\\hline
\end{tabular}

\end{minipage}

\end{table}

\subsection{Vocabulary-based analysis}
To understand the medical domain, we use the WordClouds tool\footnote{https://www.wordclouds.com/} to display visual word frequency that appears commonly in the dataset (Figure \ref{wc_ques}). Table \ref{tab:wordcloud} shows the top 10 words with the most frequency. These words are related to the medical domain. Besides, users ask many questions about Coronavirus (COVID-19), children, inflammatory diseases, and allergies.
\\
\\
\\
\begin{minipage}{0.45\linewidth}
\centering
\captionof{table}{Top 10 common words in the ViHealthQA dataset.}
\label{tab:wordcloud}
\selectlanguage{vietnamese}
\begin{tabular}{clcl}
		\toprule
\textbf{No.}        &   \textbf{Word}           & \textbf{Freq.} & \textbf{English}  \\ 
		\midrule
1&	bác sĩ &	7790&	doctor\\
2&	bé &	3409&	baby\\
3&	xét nghiệm &	3316&	test\\
4&	trẻ&	3012&	children\\
5&	triệu chứng&	2858&	symptom\\
6&	dị ứng&	2628&	allergic\\
7&	mũi &	2479&	nose\\
8&	da	&1979&	skin\\
9&	tiêm chủng &	1912&	vaccination\\
10	&gan&	1856	&liver\\
		\bottomrule

\end{tabular}
\end{minipage}\hfill
\begin{minipage}{0.45\linewidth}
\centering
\includegraphics[width=1.\textwidth, height=0.7\textwidth]{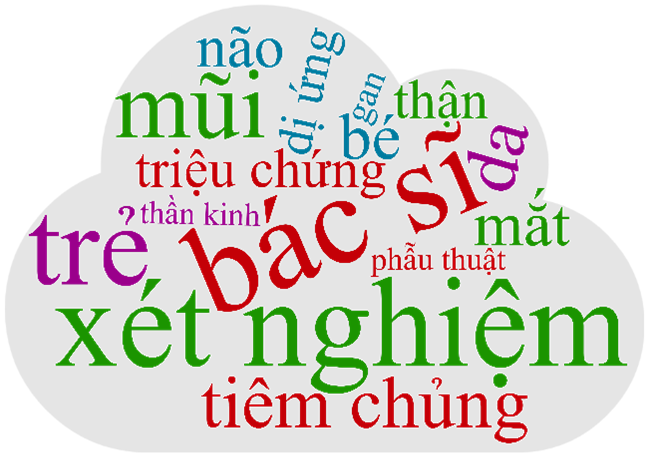}
\vspace*{5pt}
 \captionof{figure}{Word distribution of ViHealthQA.}
 \label{wc_ques}
\end{minipage}
\\

%% file: sections/5_methodologies.tex
\section{SPBERTQA: A Two-Stage Question Answering System Based on Sentence Transformers}
\begin{figure}
\centering
  \includegraphics[width=1\linewidth]{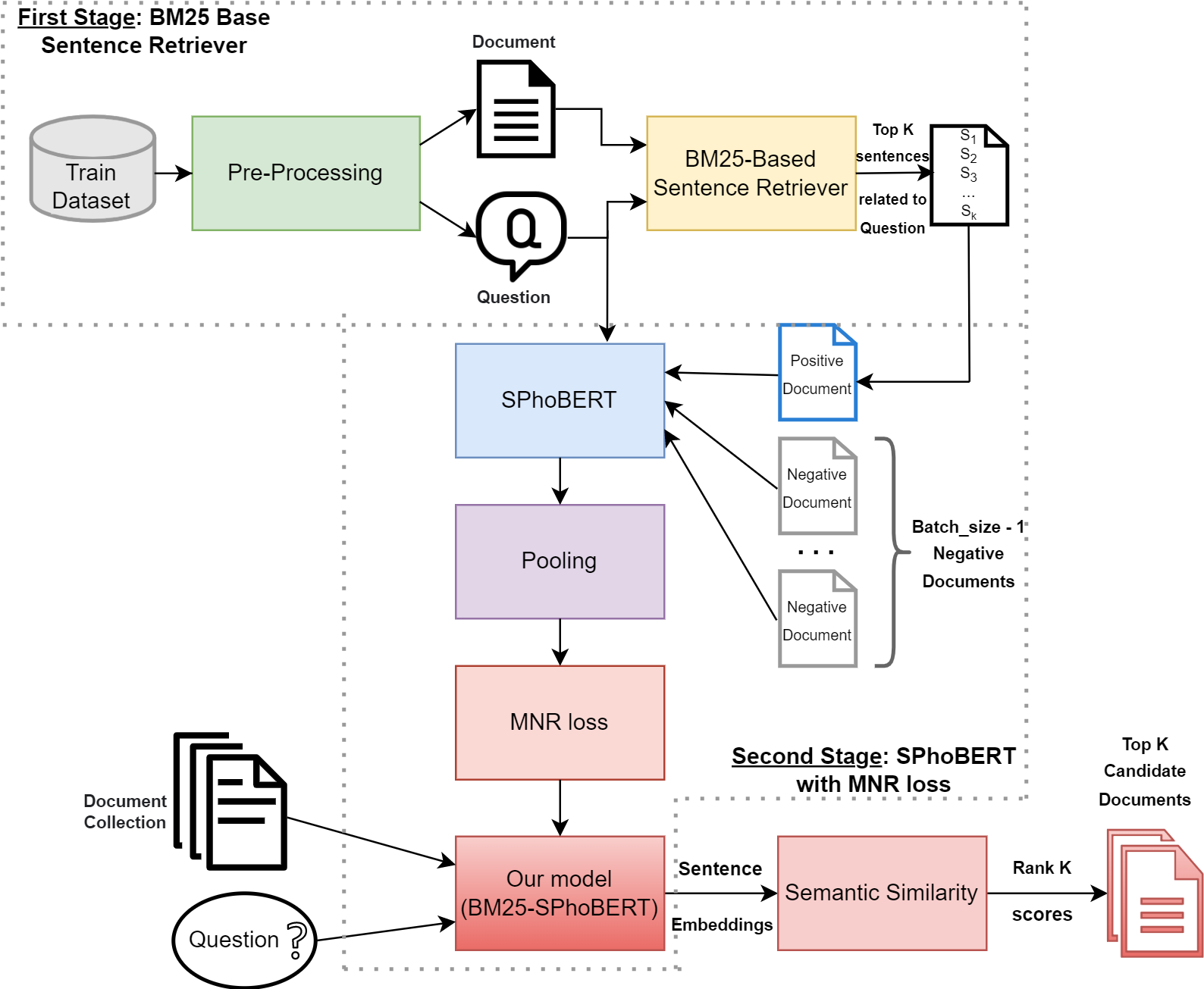}
  \caption{Overview of our system.}
  \label{system}
\end{figure}

In this paper, we propose a two-stage question answering system called SPBERTQA (Figure \ref{system}), including BM25-based sentence retriever and SBERT using PhoBERT fine-tuning with MNR loss. After training, the inputs (the question and the document collection) feed into BM25-SPhoBERT. Then, we rank the top K cosine similarity scores between sentence-embedding outputs to extract top K candidate documents.
\vspace*{-20pt}
\subsection{BM25 Based Sentence Retriever}

We aim to train the model by focusing on the meaningful knowledge of our dataset. Thus, we propose the sentence retriever stage that extracts the $K$ sentences in every answer passage the most relevant to the corresponding question. Moreover, this stage helps solve the obstacle of the maximum length sequence of every pre-trained BERT model is 512 tokens ($max\_seq\_length$ of PhoBERT = 248 tokens), while the number of answer passages over 300 tokens in Train accounts for above 65.47\%. 

We use BM25 for the first stage because BM25 mostly brings good results in IR systems \cite{Robertson2}. Besides, most answer passages have below four sentences (Average number of sentences in every answer passage $= 3.95$ in Table \ref{tab:ques_len}), so we choose $K = 5$.

\subsection{SBERT using PhoBERT and fine-tuning with MNR loss}
\textbf{Multiple negatives ranking (MNR) loss:} MNR loss works great for IR, and semantic search \cite{Henderson}. The loss function is given by Equation (1).

$$
L=-\frac{1}{N} \cdot \frac{1}{K} \cdot \sum_{i=1}^{K}\left[S\left(x_{i}, y_{i}\right)-\log \sum_{j=1}^{K}  e^{S\left(x_{i}, y_{j}\right)}\right]     (1)
$$ 
In every batch, there are $K$ positive pairs (${x_i,  y_i}$: question and positive answer passage), and each positive pair has $K - 1$ random negative answer passages ${(y_j, i\neq j)}$. The similarity between question and answer passage $(S(x, y))$ is cosine similarity. Moreover, $N$ is the Train size.

In the second stage, we use the pre-trained PhoBERT model. PhoBERT \cite{Dat} is the first public large-scale monolingual language model for Vietnamese. PhoBERT pre-training approach is based on RoBERTa, which optimizes more robust performance. Then, we fine-tune PhoBERT with MNR loss.

%% file: sections/6_experiments.tex
\section{Experiments}
\subsection{Comparative methods}
We compare our system with traditional methods such as BM25, TFIDF-Cos, and LM; pre-trained PhoBERT; and fine-tuned SBERT such as BM25-SXMLR and BM25-SmBERT.
\subsubsection{BM25} BM25 is an optimized version of TF-IDF. Equation (2) portrays the BM25 score of document ${D}$ given a query ${q}$. ${d_{avg}}$ is the length of the average document. Moreover, BM25 adds two parameters: ${k}$ helps balance the value between term frequency and ${IDF}$, and ${b}$ adjusts the importance of document length normalization. In 2008, Manning et al. \cite{Manning} suggested reasonable values are $k = [1.2,2.0]$ and $b = 0.75$.

$$
B M 25(D, q)=\underbrace{\frac{f(q, D)*(k+1)}{f(t,D)+k *\left(1-b+b * \frac{D}{d_{a v g}}\right)}}_{T F} *\underbrace{\log \left(\frac{N-N(q)+0.5}{N(q)+0.5}+1\right)}_{I D F}     \quad\quad(2)
$$
\subsubsection{TF-IDF Cosine Similarity (TFIDF-Cos)}
Cosine similarity is one of the most popular similarity measures applied to information retrieval applications and is superior to the other measures such as the Jaccard measure and Euclidean measure \cite{Subhashini}.
Given ${a}$ and ${b}$ as the respective TF-IDF bag-of-words of question and answer passage. The similarity between ${a}$ and ${b}$ is calculated by Equation (3) \cite{Pathak}. 
$$Cos(\vec{a}, \vec{b})=\frac{\vec{a} \cdot \vec{b}}{\|\vec{a}\|\|\vec{b}\|}=\frac{\sum_{1}^{n} a_{i} b_{i}}{\sqrt{\sum_{1}^{n} a_{i}^{2}} \sqrt{\sum_{1}^{n} b_{i}^{2}}}
     \quad\quad(3)$$
\subsubsection{Language Model (LM)}
LM is a probabilistic model of text \cite{Tan}. Questions and answers are modeled based on a probability distribution over sequences of words. The original and basic method for using LM is unigram query likelihood (Equation (4)).

$$
P\left(q_{i} \mid D\right)=\left(1-\alpha_{D}\right)*P\left(q_{i} \mid D\right)+\alpha_{D}*P\left(q_{i} \mid C\right)     \quad\quad(4)
$$

${P(q | D)}$ is the probability of the query q under the language model derived from ${D}$. ${P(q | C)}$ denotes a background corpus to compute unigram probabilities to avoid 0 scores \cite{Zhai}. Besides, various smoothing based on how to handle ${\alpha_{D}}$ and ${ \alpha_{D} \in [0,1]}$.

\subsubsection{PhoBERT} We directly use PhoBERT to encode question and answer passages. Then, we rank the top K answer passages having the highest cosine similarity scores with the corresponding question.
\subsubsection{BM25-SXLMR} Similar to our model, but in the second stage, we use XLM-RoBERTa instead of PhoBERT. XLM-RoBERTa \cite{Conneau} was pre-trained on 2.5TB of filtered CommonCrawl data containing 100 languages (including Vietnamese).  
\subsubsection{BM25-SmBERT} Similar to our model, but in the second stage, we use BERT multilingual. BERT multilingual was introduced by \cite{pires2019multilingual}. This model is a transformers model pre-trained on the enormous Wikipedia corpus with 104 languages (including Vietnamese) using a masked language modeling (MLM) objective.

\subsection{Data preprocessing}
We pre-process data such as lowercase, removing uninterpretable characters (e.g., new-line and extra whitespace). In order to tokenize data, we employ the RDRSegmenter of VnCoreNLP \cite{Vu}.
Moreover, stop-words can become noisy factors for traditional methods working well on pairs with high word matching between query and answer. Therefore, we conduct the removing stop-words phase. Firstly, we use TF-IDF to extract stop-words, and then we remove these words from the data. 

\subsection{Experimental settings}
We choose \textit{xlm-roberta-base}\footnote{https://huggingface.co/xlm-roberta-base}, \textit{bert-base-multilingual-cased}\footnote{https://huggingface.co/bert-base-multilingual-cased}, and \textit{vinai/phobert-base}\footnote{https://huggingface.co/vinai/phobert-base}. Then, we fine-tune SBERT with 15 epochs, batch size of 32, learning rate of $2 e^{-5}$, and maximum length of 256. 
Our experiments are performed on a single NVIDIA Tesla P100 GPU on the Google Collaboratory server\footnote{https://colab.research.google.com/}.

\subsection{Evaluation metric}
$P@K$ (Equation (5)) is the percentage of questions for which the exact answer passage appears in one of the $K$ retrieved passages \cite{van2020new}.
$$
 P@K =\frac{1}{|Q|} \sum_{1}^{n}\left\{\begin{array}{c}
1  \quad a_{q} \in A_{K}(q) \\
0\quad Otherwise 
\end{array}\right.     (5)
$$Where, ${Q = {q_1, q_2,...,q_n  }}$: collection of questions and ${q \in Q}$. ${A = {a_1, a_2,...,a_n  }}$: collection of answer passages. ${a_q}$ is exact answer-passage of question $q$. 
${A_K (q) \subseteq A}$ is the $K$ most relevant passages extracted for question $q$.

Besides, mean average precision (mAP) is used to evaluate the performance of models.

%% file: sections/7_resultsanddiscussion.tex
\section{Results and Discussion}
\subsection{Results and discussion}
With the results shown in Table \ref{pre} and \ref{tab:model}, our system achieves the best performance with 62.25\% mAP score, 50,92\% $P@1$ score, and 83.76\% $P@10$ score on the Test. BM25-SXLMR and BM25-SmBERT utilizing multilingual BERT do not work better than our system using monolingual PhoBERT. Compared to the PhoBERT model without fine-tuning with MNR loss, models fine-tuned with MNR (BM25-SXLMR, BM25-SmBERT, and our system) have good results, which proves that using MNR loss to fine-tune models for this task is suitable.
\vspace*{-10pt}
\begin{table}
\begin{minipage}{.45\linewidth}
\caption{Results on Dev and Test with $P@K$ score (\%).}
\label{pre}
\setlength{\tabcolsep}{1pt}
\begin{tabular}{lllll}
\hline
\multicolumn{1}{c}{\multirow{2}{*}{\textbf{Model}}}        & \multicolumn{2}{c}{\textbf{$P@1$}}                   & \multicolumn{2}{c}{\textbf{$P@10$}}                  \\ \cline{2-5} 
\multicolumn{1}{c}{}                                       & \multicolumn{1}{c}{Dev} & \multicolumn{1}{c}{Test} & \multicolumn{1}{c}{Dev} & \multicolumn{1}{c}{Test} \\ \hline
BM25                                                       & 51.86                   & 44.96                    & 75.93                   & 70.09                    \\
LM                                                         & 52.27                   & 47.19                    & 78.15                   & 72.38                    \\
TFIDF-Cos                                                  & 47.63                   & 39.54                    & 75.13                   & 70.39                    \\ \hline
PhoBERT                                                    & 8.36                    & 6.95                     & 31.72                   & 23.10                    \\ \hline
\begin{tabular}[c]{@{}l@{}}BM25 - SXLMR\end{tabular}    & 53.58                   & 46.05                    & 85.90                   & 79.04                    \\
\begin{tabular}[c]{@{}l@{}}BM25 - SmBERT\end{tabular} & 49.85                   & 44.91                    & 81.97                   & 75.71                    \\ \hline
\textbf{Our system}                                        & \textbf{69.52}          & \textbf{50.92}           & \textbf{89.12}          & \textbf{83.76}           \\ \hline
\end{tabular}
\end{minipage}
\hfill
\begin{minipage}{.45\linewidth}
\centering
\caption{Results on Dev and Test with mAP score (\%).}
\setlength{\tabcolsep}{5pt}
\label{tab:model}
\begin{tabular}{lll}
\hline
\multicolumn{1}{c}{\multirow{2}{*}{\textbf{Model}}} & \multicolumn{1}{c}{\multirow{2}{*}{\textbf{Dev}}} & \multicolumn{1}{c}{\multirow{2}{*}{\textbf{Test}}} \\
\\
\hline
BM25                               & 64.62                            & 56.93                             \\
LM                                 & 56.01                            & 56.00                             \\
TFIDF-Cos                          & 57.12                            & 50.31                             \\ \hline
PhoBERT                            & 16.08                            & 12.45                             \\ \hline
BM25-SXLMR                         & 59.96                            & 53.85                             \\
BM25-SmBERT                      & 60.77                            & 55.52                             \\ \hline
\textbf{Our system}                & \textbf{69.52}                   & \textbf{62.25}                    \\ \hline
\end{tabular}
\end{minipage}
\end{table}
\vspace*{-20pt}

\subsection{Analysis}
\begin{figure}[!hbt]
\centering
\includegraphics[width=0.9\linewidth]{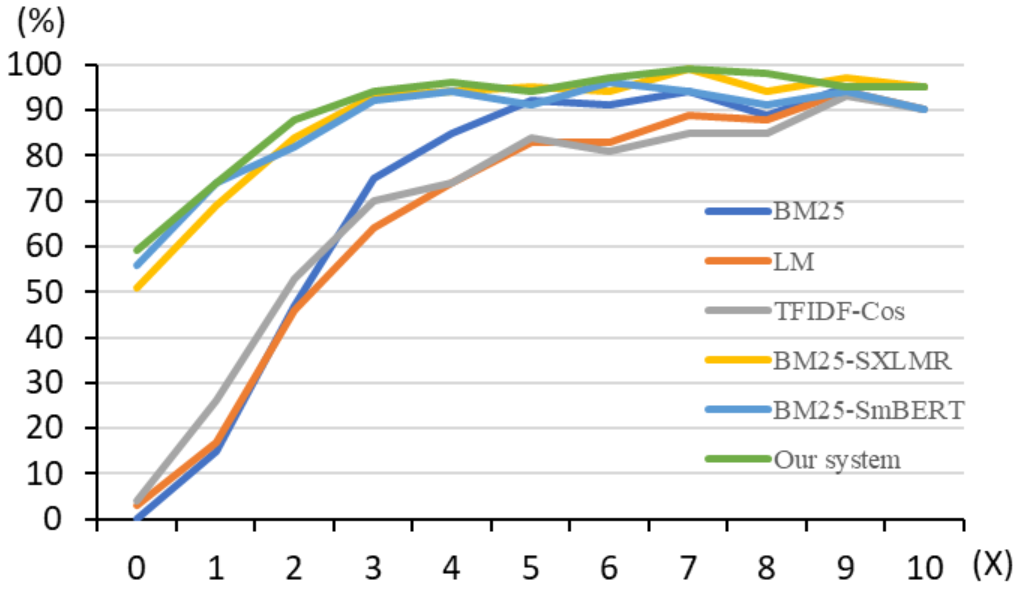}
\caption{Results of lexical overlap experiments with P@1 (\%).}
\label{overlap}
\end{figure}
To understand deeply about our system is more robust than traditional methods, and traditional methods have disadvantages in lexical gap issues, we run models on pairs having lexical overlap (the number of duplicate words between question and answer passage – $X$) from 0 to 10. As results are shown in Figure \ref{overlap}, with $X < 4$, bag-of-words methods cannot extract the precise answer. Especially with $X = 0$, these models mostly do not work. While, with $X = 0$, fine-tuned models have results with an upper 50\% $P@1$ score. From $K = 3$, these models have good scores with an upper 80\% $P@1$ score. 
Moreover, we provide typical examples of Dev predicted by BM25, LM, and our system (Table \ref{ana}). ID 169 has word matching between question and answer passage. The models that can retrieve precise answers are BM25, LM, and our system. In contrast, in ID 776, no words of question appear in the answer passage. Hence, the models must understand the semantic backgrounds instead of capturing high lexical overlap information to retrieve the precise answer. BERT models capture context and meaning better than bag-of-words methods \cite{Han}. In particular, SBERT can derive semantically meaningful sentence embeddings \cite{Reimers}. Therefore, our system based on sentence transformers can find the exact answer passage for the question with ID 776.
\begin{table}[!hbt]
\centering
\caption{Examples in Dev predicted by traditional methods and our system.}
\label{ana}
\selectlanguage{vietnamese}
\begin{tabular}{cp{0.3\textwidth}p{0.6\textwidth}p{0.13\textwidth}}
  \toprule
  \textbf{ID} & \centering\textbf{Question}& \centering\textbf{Answer passage}&\textbf{Models} \\
  \midrule
  776   & Tai biến, chân tay tê bì điều trị như thế nào? (\emph{How is the stroke and tingling in in hands and feet treated?})&Nếu vấn đề chính là rối loạn điện giải, nhiễm trùng huyết và gan thận, bạn nên đưa bố đến khám chuyên khoa Nội tiết hoặc Nội tổng quát. Về thần kinh, bác sĩ khám cần xem lại phim CT/MRI não để đánh giá lại tổn thương não mới có thể có được kế hoạch phòng ngừa đột quỵ tái phát, điều trị giảm đau thần kinh và phục hồi chức năng tối ưu.
(\emph{If the main problem is electrolyte disturbances, sepsis, and hepatobiliary disease, you should take him to see an Endocrinologist or General Internal Medicine. Neurologically, the examining doctor needs to review the brain CT/MRI film to re-evaluate the brain damage so that he can have a plan to prevent recurrent stroke, treat neuropathic pain, and restore optimal function.})& Our system
 \\
  \midrule
169   & \textbf{Bệnh suy tủy xương vô căn} có nguy hiểm không và \textbf{điều trị} thế nào?
(\emph{Is \textbf{bone marrow failure syndromes} dangerous and how is \textbf{treatment}?})&\textbf{Suy tủy xương vô căn} tùy thuộc vào từng giai đoạn thì cách \textbf{điều trị} khác nhau. Nếu số lượng máu quá thấp thì phải điều trị ức chế miễn dịch hoặc ghép tủy. Có những bệnh nhân không đáp ứng với \textbf{điều trị}, tuy nhiên cũng có nhiều bệnh nhân chữa khỏi.
(\emph{\textbf{Bone marrow failure syndromes} depends on the stage, the \textbf{treatment} is different. If the blood count is too low, then immunosuppressive therapy or bone marrow transplant is required. Some patients do not respond to \textbf{treatment}, but many patients are cured.})
&BM25, LM, and our system
\\
  \bottomrule
\end{tabular}
\end{table}

%% file: sections/8_conclusion.tex
\section{Conclusion and Future Work}

In this paper, we created the ViHealthQA dataset that comprises 10,015 question-answer passage pairs in the medical domain. Every answer passage is a doctor's reply to the corresponding user's question, so the ViHealthQA dataset is suitable for real search engines. Secondly, we propose the SPBERTQA, a two-stage question answering system based on sentence transformers on our dataset. Our proposed system performs best over bag-of-word-based models and fine-tuned multilingual pre-trained language models. This system solves the problem of linguistic gaps.

In future, we plan to employ the machine reading comprehension (MRC) module. This module helps extract answer spans from answer passages so that users can comprehend the meaning of the answer faster.